\documentclass[runningheads]{llncs}

\usepackage{eccv}

\usepackage{eccvabbrv}
\usepackage{graphicx}
\usepackage{booktabs}

\usepackage[accsupp]{axessibility}  

\usepackage{hyperref}

\usepackage{orcidlink}

\begin{document}

\title{Champ: Controllable and Consistent Human Image Animation with 3D Parametric Guidance} 

\titlerunning{Champ: Controllable Human Image Animation w. 3D Parametric Guidance}

\author{Shenhao Zhu*\inst{1} \and
Junming Leo Chen*\inst{2} \and
Zuozhuo Dai\inst{3} \and
Qingkun Su\inst{3} \and \\
Yinghui Xu\inst{2} \and
Xun Cao\inst{1} \and 
Yao Yao\inst{1} \and
Hao Zhu$^\dag$\inst{1} \and
Siyu Zhu$^\dag$\inst{2} }

\authorrunning{S. Zhu et al.}

\institute{Nanjing University, Nanjing, China \and
Fudan University, Shanghai, China \and
Alibaba Group, Hangzhou, China \\
\email{shenhaozhu@smail.nju.edu.cn, leochenjm@gmail.com, \\ \{caoxun, yaoyao, zh\}@nju.edu.cn, zuozhuo.dzz@alibaba-inc.com, \\ suqingkun@gmail.com, \{xuyinghui, siyuzhu\}@fudan.edu.cn}}

\maketitle

\begin{abstract}
In this study, we introduce a methodology for human image animation by leveraging a 3D human parametric model within a latent diffusion framework to enhance shape alignment and motion guidance in curernt human generative techniques.
The methodology utilizes the SMPL(Skinned Multi-Person Linear) model as the 3D human parametric model to establish a unified representation of body shape and pose. 
This facilitates the accurate capture of intricate human geometry and motion characteristics from source videos.
Specifically, we incorporate rendered depth images, normal maps, and semantic maps obtained from SMPL sequences, alongside skeleton-based motion guidance, to enrich the conditions to the latent diffusion model with comprehensive 3D shape and detailed pose attributes.
A multi-layer motion fusion module, integrating self-attention mechanisms, is employed to fuse the shape and motion latent representations in the spatial domain.
By representing the 3D human parametric model as the motion guidance, we can perform parametric shape alignment of the human body between the reference image and the source video motion. 
Experimental evaluations conducted on benchmark datasets demonstrate the methodology's superior ability to generate high-quality human animations that accurately capture both pose and shape variations. 
Furthermore, our approach also exhibits superior generalization capabilities on the proposed in-the-wild dataset. Project page: \url{https://fudan-generative-vision.github.io/champ}.

\keywords{Latent Diffusion Model \and Human Image Animation \and 3D human parametric model \and Motion Guidance}
\end{abstract}
\let\thefootnote\relax\footnotetext{$^*$ These authors contributed equally to this work.}
\let\thefootnote\relax\footnotetext{ $^\dag$ Corresponding Author}
\begin{figure}[!t]
  \centering
  \includegraphics[width=0.98\linewidth]{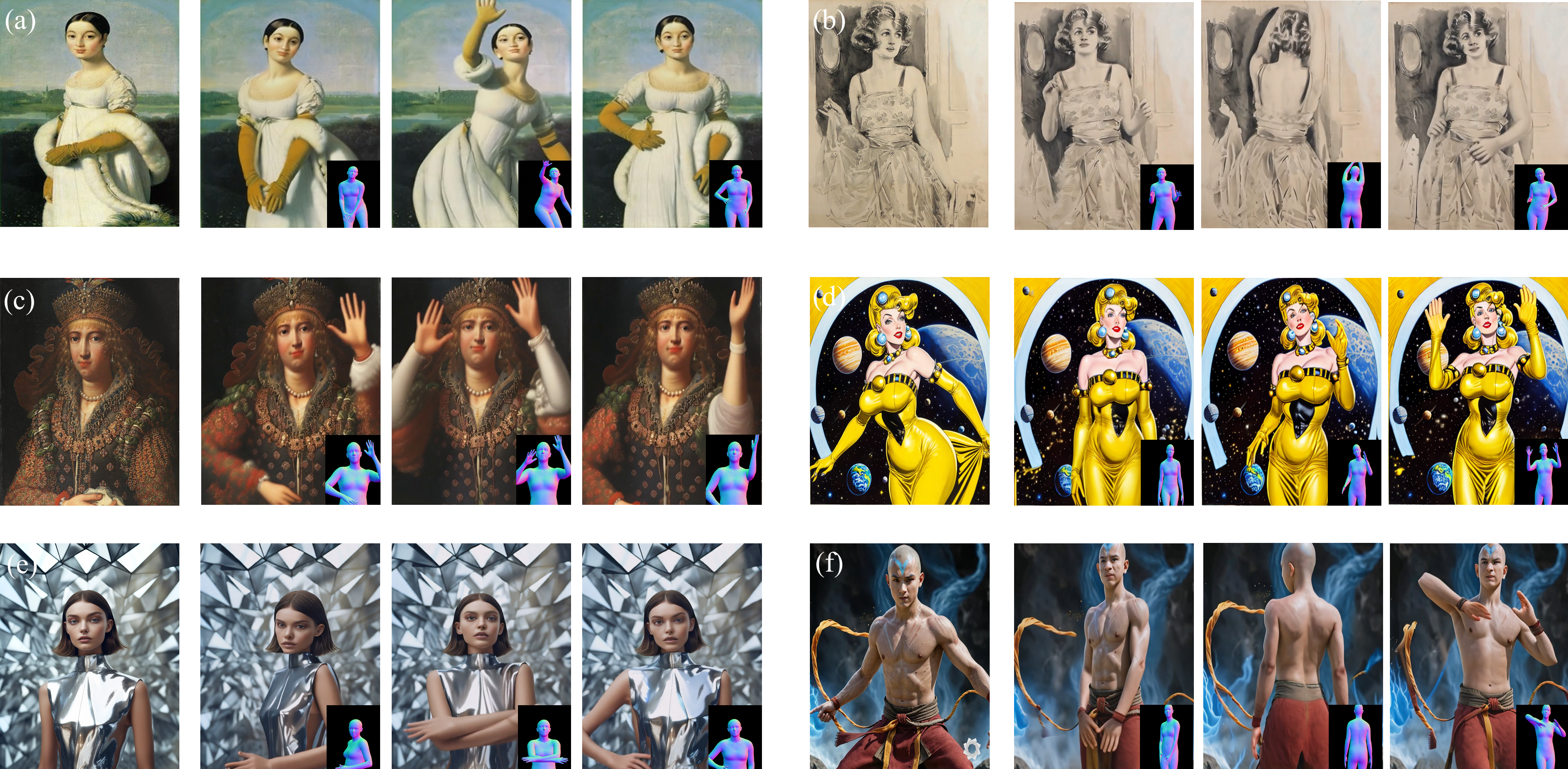}
  \caption{
The proposed methodology showcases a novel ability to produce temporally consistent and visually authentic human image animations by leveraging a reference image and a prescribed sequence of motion articulated through 3D human parametric models. 
Furthermore, it demonstrates an enhanced capacity to refine shape alignment and motion guidance within the resulting videos.
This approach facilitates the animation of a wide range of characters, encompassing portraits exhibiting substantial domain variations, such as: 
(a) a neoclassical oil painting portraying a woman adorned in a white dress and fur coat;
(b) a watercolor portrait of a woman;
(c) an oil panel painting titled ``The Queen of Armenia'', as well as characters derived from Text-to-Image diffusion models with the following prompts: 
(d) a painting of a woman in a yellow dress, heavy metal comic cover art, space theme; 
(e) a woman in a silver dress posing for a picture, trending on cg society, futurism, with bright blue eyes; 
(f) a realistic depiction of Aang, the last airbender, showcasing his mastery of all bending elements while in the powerful Avatar State.}
\vspace{-5mm}
\label{fig:teaser}
\end{figure}

\section{Introduction}
\label{sec:intro}
Recent advancements in generative diffusion models, particularly latent diffusion models, have significantly propelled the field of image animation forward~\cite{yu2023bidirectionally,zhang2023adding,zhao2022thin,li2024synthesizing}. These advancements have found broad application in virtual reality experiences, interactive storytelling, and digital content creation, resulting in the production of a plethora of sophisticated dynamic visual content.
Within the realm of human image animation, techniques typically rely on a reference image and motion guidance specific to humans, such as skeletons~\cite{wang2023disco,hu2023animate}, semantic maps~\cite{siarohin2021motion,xu2023magicanimate}, and dense motion flows~\cite{zhao2022thin,karras2023dreampose}, to generate controllable human animation videos.
In this domain, two predominant approaches prevail: those based on GANs~\cite{goodfellow2014generative,mirza2014conditional} and diffusion models~\cite{ho2020denoising,song2020score}.

GAN based methods~\cite{siarohin2019first,wang2021one,tian2021good,wang2020g3an, yoon2021poseguided, sarkar2021humangan} commonly employ warping functions to spatially transform the reference image according to input motion for generating sequential video frames. 
By leveraging the inherent generative visual capabilities of GANs, these methods aim to fill in missing regions and improve visually implausible areas within the generated content.
Despite yielding promising results in dynamic visual content generation, GAN-based approaches often encounter challenges in effectively transferring motion, particularly in scenarios involving substantial variations in human identity and scene dynamics between the reference image and the source video motion. 
This limitation manifests as unrealistic visual artifacts and temporal inconsistencies in the synthesized content.

Simultaneously, diffusion-based methodologies~\cite{guo2023animatediff,karras2023dreampose,wang2023disco,karras2023dreampose,zhang2023magicavatar} incorporate the reference image and various dynamics as conditions at both the appearance and motion levels.
By harnessing the generative capabilities of latent diffusion models in conjunction with condition guidance, these techniques facilitate the direct generation of human animation videos. 
Recent diffusion models~\cite{karras2023dreampose,wang2023disco}, grounded in data-driven strategies, notably those leveraging CLIP-encoded visual features~\cite{radford2021learning} extracted from reference images pretrained on a vast collection of image-text pairs, in conjunction with diffusion models and temporal alignment modules, have demonstrated effectiveness in addressing the generalization challenges inherent in GAN-based approaches.
Therefore, inspired by advanced methodologies such as Animate Anyone~\cite{hu2023animate} and MagicAnimate~\cite{xu2023magicanimate}, the aim of this paper is to further optimize shape alignment and pose guidance mechanisms.

In the present study, we propose that the use of a reference image in conjunction with pose guidance, provided through sequential skeleton or dense pose data, may present certain limitations concerning both pose alignment and motion guidance.
In a progressive step forward, we advocate for the adoption of a 3D parametric human model, such as SMPL~\cite{SMPL:2015}, to encode the 3D geometry of the reference image and extract human motion from the source videos.
Firstly, diverging from approaches that segregate the representation of body shape and pose (e.g. dense pose and skeleton methods that primarily emphasize pose), the SMPL model offers a unified representation that encompasses both shape and pose variations using a low-dimensional parameter space.
Consequently, in addition to pose information, SMPL model also provides guidance on human geometry-related surface deformations, spatial relationships (e.g. occlusions), contours, and other shape-related features.
Secondly, owing to the parametric nature of the SMPL model, we can establish geometric correspondence between the reconstructed SMPL from the reference image and the SMPL-based motion sequences extracted from the source video. 
This enables us to adjust the parametric SMPL-based motion sequences, thereby enhancing the motion and geometric shape conditions within latent diffusion models.
Thanks to the generalization of SMPL to different body shapes, we can effectively handle the substantial variations in body shapes between the reference image and source video.

Incorporating the SMPL model as a guiding framework for both shape and pose within a latent diffusion model, our methodology is structured around three fundamental components:
1) The sequences derived from the SMPL model corresponding to the source video are projected onto the image space, resulting in the generation of depth images, normal maps, and semantic maps that encapsulate essential 3D information. 
The depth image plays a critical role in capturing the 3D structure of the human form, while the normal map enables the representation of orientation of the human body. 
The semantic map aids in accurately managing interactions between different components of the human body during the process of animation generation.
2) Our analysis demonstrates that incorporating skeleton-based motion guidance enhances the precision of guidance information, particularly for intricate movements such as facial expressions and finger movements. 
As a result, the skeleton is maintained as an auxiliary input to complement the aforementioned maps.
3) In the process of integrating depth, normal, semantic, and skeleton maps through feature encoding, the employment of self-attention mechanisms facilitates the feature maps, processed via self-attention, in learning the representative saliency regions within their respective layers. 
Such multi-layer semantic fusion enhances the model's capability to comprehend and generate human postures and shapes. 
Finally, the inclusion of these multi-layer feature embeddings conditioned on a latent video diffusion model leads to precise image animation both in pose and shape.

Our proposed methodology has been evaluated through thorough experiments using the popular TikTok and UBC fashion video datasets, showcasing its effectiveness in improving the quality of human image animation.
Furthermore, we have conducted a comparative analysis of our methodology against state-of-the-art approaches on a novel video dataset gathered from diverse real-world scenarios, demonstrating the robust generalization capabilities of our proposed approach.
    

\section{Related Work}
\label{sec:related_work}

\textbf{Diffusion Models for Image Generation.}
Diffusion-based models~\cite{balaji2022ediffi,huang2023composer,nichol2021glide,ramesh2022hierarchical,rombach2022high,saharia2022photorealistic} have rapidly emerged as a fundamental component in the domain of text-to-image generation, renowned for their capacity to yield highly promising generative outcomes. 
To address the considerable computational requirements inherent in diffusion models, the Latent Diffusion Model, as proposed in~\cite{rombach2022high} introduces a technique for denoising within the latent space.
This method not only enhances the computational efficiency of these models but also preserves their ability to generate high-fidelity images.
Moreover, in the endeavor to enhance control over visual generation, recent studies such as ControlNet~\cite{zhang2023adding}, T2I-Adapter~\cite{mou2023t2i}, and IP-Adapter~\cite{ye2023ip} have delved into the incorporation of supplementary encoder layers.
These layers facilitate the assimilation of control signals encompassing aspects such as pose, depth, and edge information, and even permit the utilization of images in conjunction with textual prompts.
This progression signifies a significant advancement towards more controlled and precise image generation, facilitating the creation of images characterized by not only superior quality but also enriched contextual accuracy and detail.

\textbf{Diffusion Models for Human Image Animation.}
The task of animating human images, a significant endeavor within the domain of video generation, aims to seamlessly create videos from one or multiple static images~\cite{chan2019everybody,ren2020deep,siarohin2019first,siarohin2021motion,yu2023bidirectionally,zhang2022exploring,zhao2022thin,yoon2021pose, sarkar2021neural, hu2023sherf, albahar2023humansgd, cao2023dreamavatar, prokudin2021smplpix, fu2022styleganhuman, jiang2023humangen}.
The recent advancements of diffusion models in the text-to-image domain have sparked interest in exploring their utility for animating human images.
PIDM~\cite{bhunia2023person} introduces a texture diffusion module that is specifically crafted to align the texture patterns of the source and target images closely, thereby enhancing the realism of the resultant animated output.
DreamPose~\cite{karras2023dreampose} capitalizes on the capabilities of the pre-trained Stable Diffusion model by incorporating both CLIP~\cite{radford2021learning} and VAE~\cite{kingma2013auto} for image encoding. 
It integrates these embeddings with an adapter. 
Similarly, DisCo~\cite{wang2023disco} innovatively segregates the control of pose and background using dual independent ControlNets~\cite{zhang2023adding}, providing finer control over the animation process. 
Animate Anyone~\cite{hu2023animate} utilizes a UNet-based ReferenceNet to extract features from reference images.
It includes pose information via a lightweight pose guider. Expanding on the principles introduced by AnimateDiff~\cite{guo2023animatediff}, Animate Anyone integrates a temporal layer into the denoising UNet to enhance temporal coherence.
MagicAnimate~\cite{xu2023magicanimate} follows a similar approach but employs a ControlNet tailored for DensePose \cite{guler2018dense} inputs instead of the more commonly used OpenPose~\cite{cao2017realtime} keypoints to provide more precise pose guidance.
This paper primarily builds upon esteemed diffusion-based methodologies and advances the optimization of appearance alignment and motion guidance mechanisms. 
This is achieved by introducing a 3D parametric model for geometric reconstruction of the reference image and motion modeling of the source video sequence.

\textbf{Pose Guidance in Human Image Animation.}
DWpose\cite{yang2023effective} stands out as an enhanced alternative to OpenPose\cite{cao2017realtime}, offering more accurate and expressive skeletons. 
This improvement has proven beneficial for diffusion models in generating higher quality images, with its adoption as a condition signal in various works\cite{feng2023dreamoving,hu2023animate}.
The work presented in DensePose~\cite{Guler2018DensePose} aims to establish dense correspondences between an RGB image and a surface-based representation.
The SMPL~\cite{SMPL:2015} model is a 3D model renowned for its realistic depiction of human bodies through skinning and blend shapes.
Its widespread adoption spans fields like human reconstruction\cite{he2021arch,alldieck2018video} and interaction with environments\cite{hassan2021populating,ma2020learning}. 
It also serves as essential ground truth for neural networks in pose and shape analysis\cite{lu2023dposer,mu2023actorsnerf}.
In this paper, we consider SMPL, the 3D parametric model, to reconstruct the poses as well as the shapes from the source video, and obtain more complete condition for appearance alignment and pose guidance.

\begin{figure}[t]
  \centering
  \includegraphics[width=0.95\linewidth]{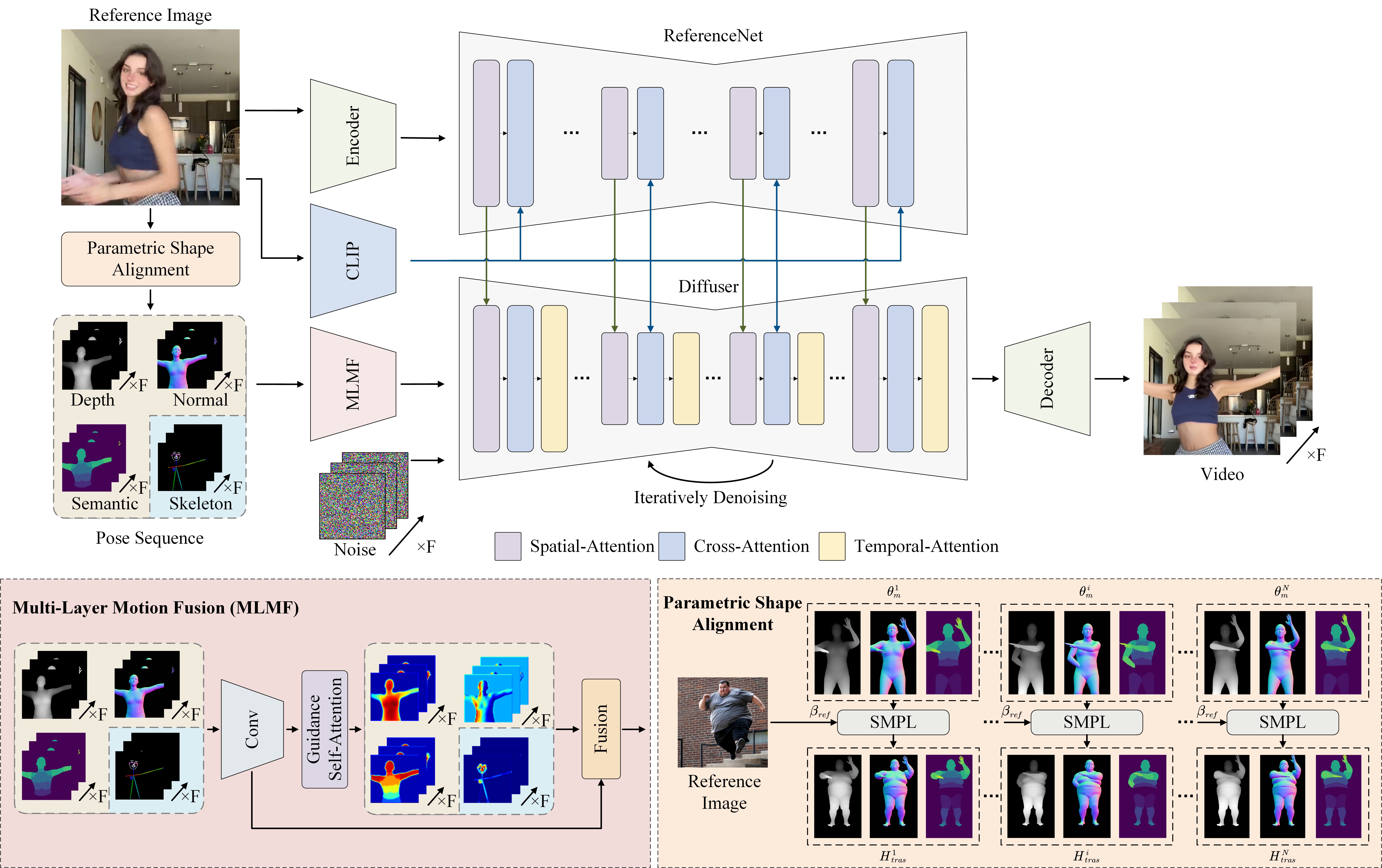}
  \caption{The overview of our proposed approach. Given an input human image and a reference video depicting a motion sequence. We obtain the pose sequence corresponding to the reference image through Parametric Shape Alignment as 3D motion guidance. MLMF is employed to encode multi-layer 3D-related motion information. Referencenet and Temporal-attention ensure identity consistency and temporal coherence, respectively.}
  \vspace{-6mm}
  \label{fig:network}
\end{figure}
\section{Method}
\label{sec:method}

Figure~\ref{fig:network} illustrates the overview of our proposed approach.
Given an input human image and a reference video depicting a motion sequence, the objective is to synthesize a video where the person in the image replicates the actions observed in the reference video, thereby creating a controllable and temporally coherent visual output.
In Section~\ref{subsec:preliminary}, we present an overview of the latent diffusion model and the SMPL model to establish the foundational concepts necessary for the subsequent discussions.
Section~\ref{subsec:condition} elaborates on the application of the SMPL model to extract pose and shape information from the source video, enabling the generation of multiple outputs containing pose and shape details.
In Section~\ref{subsec:guidance}, these outputs are then utilized to provide multi-layer pose and shape guidance for the human image animation within the latent diffusion model framework.
Lastly, Section~\ref{subsec:network} provides a comprehensive exposition of the network architecture along with a detailed description of the training and inference procedures employed in the proposed methodology.

\subsection{Preliminary}
\label{subsec:preliminary}

\textbf{Latent Diffusion Models.}
The Latent Diffusion Model (LDM) proposed by Rombach et al.~\cite{rombach2022high} presents a novel approach in the domain of Diffusion Models by incorporating two distinct stochastic processes, namely diffusion and denoising, into the latent space.
Initially, a Variational Autoencoder (VAE)~\cite{kingma2013auto} is trained to encode the input image into a low-dimensional feature space. 
Subsequently, the input image $I$ is transformed into a latent representation $\boldsymbol{z}_0 = \mathcal{E}(I)$ using a frozen encoder $\mathcal{E}(\cdot)$.
The diffusion process involves applying a variance-preserving Markov process~\cite{sohl2015deep,ho2020denoising,song2020score} to $\boldsymbol{z}_0$, where noise levels increase monotonically to generate diverse noisy latent representations:
\begin{equation}
    \label{eq:diffusion_forward}
    \boldsymbol{z}_t = \sqrt{\Bar{\alpha}_t}\boldsymbol{z}_0 + \sqrt{1- \Bar{\alpha}_t}\boldsymbol{\epsilon}, \quad \epsilon \sim \mathcal{N}(\boldsymbol{0},\boldsymbol{I})
\end{equation}
Here, $t = {1, ..., T}$ signifies the time steps within the Markov process, where $T$ is commonly configured to 1000, and $\overline{\alpha}_t$ represents the noise intensity at each time step. Subsequent to the ultimate diffusion iteration, $q(\boldsymbol{z}_T \mid \boldsymbol{z}_0)$ approximates a Gaussian distribution $\mathcal{N}(\boldsymbol{0},\boldsymbol{I})$.

The denoising process involves the prediction of noise $\boldsymbol{\epsilon}_{\theta}(\boldsymbol{z}_t,t,\boldsymbol{c})$ for each timestep $t$ from $\boldsymbol{z}_t$ to $\boldsymbol{z}_{t-1}$.
Here, $\boldsymbol{\epsilon}_{\theta}$ denotes the noise prediction neural networks, exemplified by architectures like the U-Net model~\cite{ronneberger2015u}, while $c_{\text{text}}$ signifies the text embedding derived from the CLIP mechanism.
The loss function quantifies the expected mean squared error (MSE) between the actual noise $\boldsymbol{\epsilon}$ and the predicted noise $\boldsymbol{\epsilon}_{\theta}$ conditioned on timestep $t$ and noise $\boldsymbol{\epsilon}$:
\begin{equation}
    L = \mathbb{E}_{\mathcal{E}(I), c_{\text{text}},\epsilon\sim\mathcal{N}(0,1),t}\left[\omega(t) \lVert \epsilon-\epsilon_{\theta}(z_t, t, c_{\text{text}}) \rVert_{2}^{2} \right], t=1,...,T
\end{equation}
Here, $\omega(t)$ represents a hyperparameter that governs the weighting of the loss at timestep $t$. 
Following training, the model is capable of progressively denoising from an initial state $\boldsymbol{z}_T \sim \mathcal{N}(\boldsymbol{0},\boldsymbol{I})$ to $\boldsymbol{z}_0$ using a fast diffusion sampler~\cite{song2020denoising, lu2022dpm}.
Subsequently, the denoised $\boldsymbol{z}_0$ is decoded back into the image space $I$ utilizing a frozen decoder $\mathcal{D}(\cdot)$

\textbf{SMPL model.} 
The SMPL model, as introduced in the work by Loper et al.~\cite{SMPL:2015}, stands as a prevalent methodology within the domains of computer graphics and computer vision, particularly in the realm of realistic human body modeling and animation. 
This model is structured around a parametric shape space that effectively captures the nuanced variations in body shape exhibited among individuals, alongside a pose space that intricately encodes the articulation of the human body. 
Through the amalgamation of these two spaces, the SMPL model exhibits the capability to produce anatomically plausible and visually realistic human body deformations spanning a diverse spectrum of shapes and poses.
The SMPL model operates on low-dimensional parameters for pose, denoted as $\theta \in \mathbb{R}^{24 \times 3 \times 3}$, and shape, denoted as $\beta \in \mathbb{R}^{10}$.
By inputting these parameters, the model generates a 3D mesh representation denoted as $M \in \mathbb{R}^{3 \times N}$ with $N = 6890$ vertices. 
A vertex-wise weight $W \in \mathbb{R}^{N \times k}$ is applied to evaluate the relations between the vertex and the body joints $J \in \mathbb{R}^{3 \times k}$, which could then be used for human part segmentation.

\subsection{Multi-Layer Motion Condition}
\label{subsec:condition}

\textbf{SMPL to Guidance Conditions}. 
Given a reference human image $I_\text{ref}$ and a sequence of driving motion video frames $I^{1:N}$, where $N$ denotes the total number of frames, we obtain the 3D human parametric SMPL model, $H_\text{ref}$ and $H_\text{m}^{1:N}$,  respectively, utilizing an existing framework known as 4D-Humans~\cite{goel2023humans}.
In order to extract comprehensive visual information from the pixel space, we render the SMPL mesh to obtain 2D representations. 
This includes encoding depth maps, which contain distance information from each pixel to the camera, crucial for reconstructing the 3D structure of the scene. 
Additionally, we encode normal maps, which depict the surface orientation at each point in the image and can capture geometric information related to the orientation of the human body surface. 
Furthermore, semantic segmentation maps provide class information for each pixel in the image, enabling accurate handling of interactions between different components of the human body.

\textbf{Parametric Shape Alignment.}
As a key of human video generation, animating the reference human image by driving motion sequence while keeping the reference appearance and shape remains challenging. 
Previous skeleton-based methods use only sparse keypoints to guide the animation and thus ignore the shape variety of humans. 
With the parametric human model, our work is easy to align both shape and pose between reference human and motion sequence. 
Given a SMPL model $H_\text{ref}$ fitted on reference image $I_\text{ref}$ and a SMPL sequence $H_\text{m}^{1:N}$ from N frames driving video $I^{1:N}$, we aim to align the shape $\beta_\text{ref}$ of  $H_\text{ref}$ to the pose sequence $\theta_\text{m}^{1:N}$ of $H_\text{m}^{1:N}$. The aligned SMPL model of each frame $i \in [1, N] $ is then formulated as:
 \begin{equation}
    H_\text{trans}^{i} = \text{SMPL}(\beta_\text{ref}, \theta_\text{m}^{i} )
\end{equation}
We then take corresponding conditions rendered from the $H_\text{trans}^{1:N}$ to guide video generation on image $I_\text{ref}$, which produces pixel-level aligned human shape and enhances the human appearance mapping process in generated animation video.

\subsection{Multi-Layer Motion Guidance}
\label{subsec:guidance}

\begin{figure}[t]
  \centering
  \includegraphics[width=1.0\linewidth]{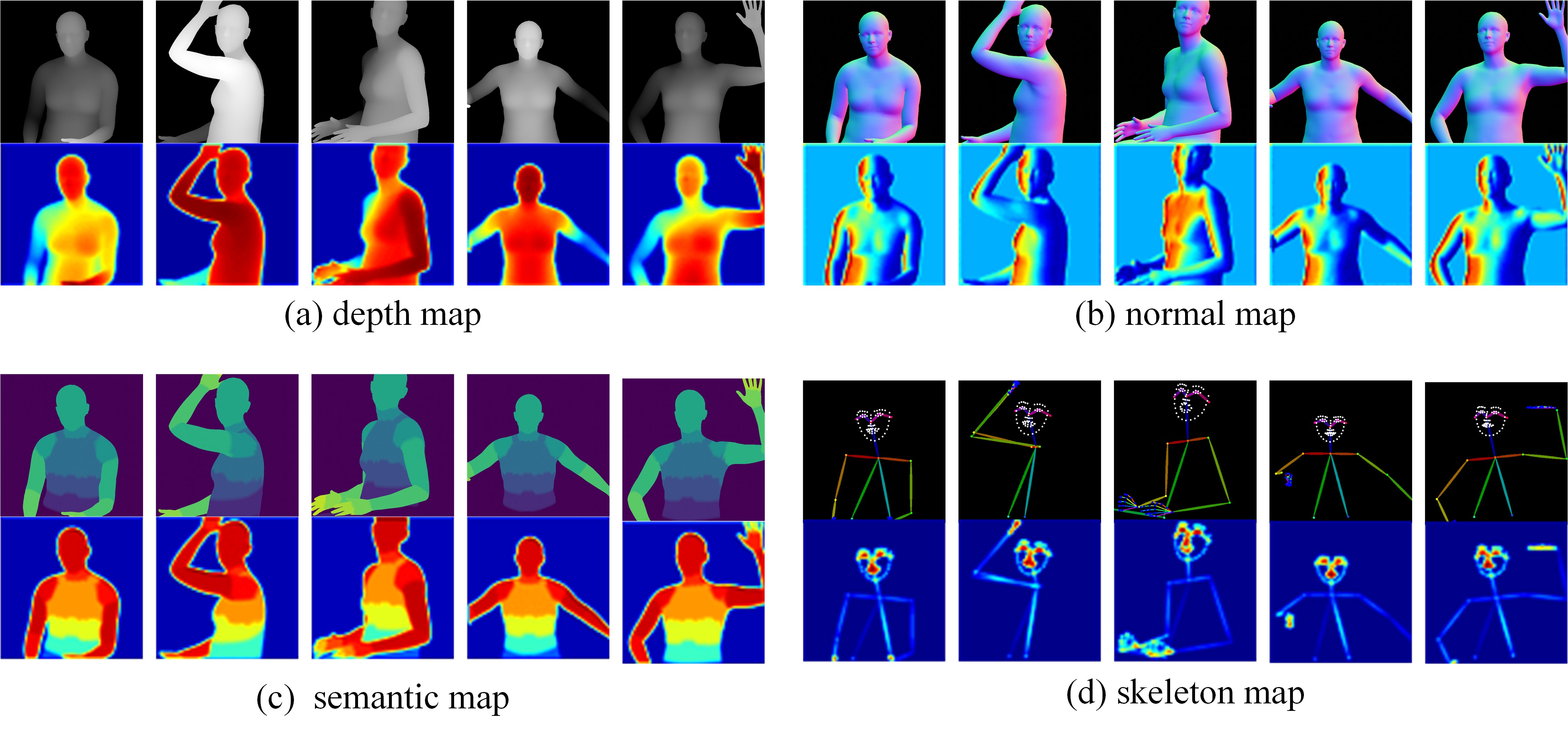}
  \caption{Multi-layer motion condition and corresponding cross attention maps. 
Each set of images (above) comprises representations of a depth map, normal map, semantic map, and DWpose skeleton rendered from the corresponding SMPL sequences.
The subsequent images (below) illustrate the output of the guidance self-attention.}
  \vspace{-5mm}
  \label{fig:motion_guidance}
\end{figure}

Now we have completed the shape-level alignment between the parametric SMPL model reconstructed based on the reference image and the SMPL model sequence of the source video using parametric shape alignment.
Subsequently, depth maps, normal maps, and semantic maps are rendered from the aligned SMPL model sequence.
Additionally, a skeleton was introduced as an auxiliary input to enhance the representation of intricate movements, such as facial expressions and finger movements. 
As shown in Figure~\ref{fig:motion_guidance}, leveraging latent feature embedding and the self-attention mechanism to be introduced below, we can spatially weight the multi-layer embeddings of human shapes and poses, resulting in the generation of multi-layer semantic fusion as the motion guidance.

\textbf{Guidance Self-Attention.}
ControlNet~\cite{zhang2023adding} is frequently used in human animation tasks to control generated actions considering additional guidance. 
However, introducing multiple guidance condition to ControlNet would result in a computational burden that is unaffordable.
In light of this, we are inspired by the advanced work~\cite{hu2023animate} and propose a guidance encoder designed to encode our multilevel guidance. 
Through this approach, we achieve the simultaneous extraction of information from the guidance while fine-tuning a pre-trained denoising U-Net.
The encoder consists of a series of lightweight networks. We assign a guidance network to each guidance condition to encode its features. 
For each guidance network, we first extract features of the guidance condition through a set of convolutional layers. 
Considering the presence of multilevel guidance conditions, which involve different characteristics of the human body, a self-attention module is appended after the convolutional layers.
This module facilitates the precise capture of corresponding semantic information for each of the multi-layer guidance condition. 
In particular, Figure~\ref{fig:motion_guidance} illustrates the self-attention map of depth, normal, semantic, and skeleton feature embeddings post-training.
The analysis reveals distinct patterns: the depth condition predominantly focuses on the geometric contours of the human figure; 
the normal condition emphasizes the orientation of the human body; 
the semantic condition prioritizes the semantic information of different body parts; 
and the skeleton attention provides detailed constraints on the face and hands.

\textbf{Multi-Layer Motion Fusion.}
In order to preserve the integrity of the pretrained denoising U-Net model, we opt to use a convolutional layer with zero initialization as the output layer to extract the features of each guidance condition.
The guidance encoder consolidates the feature embeddings from all the guidance conditions by aggregating them through summation, yielding the ultimate guidance feature denoted as $y$. 
This operation can be expressed mathematically as: 
\begin{equation}
    y = \sum_{i=1}^{N}{\mathcal{F}^i(\cdot, \theta^i)},
\end{equation}
where $N$ signifies the total count of guidance conditions incorporated, $i$ is the index of the pose guidance, and $\theta$ is the input pose image. 
Subsequently, the guidance feature is combined with the noisy latent representation before being fed into the denoising fusion module.

\subsection{Network}
\label{subsec:network}



\textbf{Network Structure.}
In this section, we present the comprehensive pipeline of our proposed method illustrated in Figure~\ref{fig:network}. 
Our approach introduces a video diffusion model that incorporates motion guidance derived from 3D human parametric models.
Specifically, we employ the SMPL model to extract a continuous sequence of SMPL poses from the motion data. 
This conversion results in a multilevel guidance that encapsulates both 2D and 3D characteristics, thereby enhancing the model's comprehension of human shape and pose attributes. 
To integrate this guidance effectively, we introduce a motion embedding module that incorporates the multilayer guidance into the model.
The multiple latent embeddings of motion guidance are individually refined through self-attention mechanisms and subsequently fused together using a multi-layer motion fusion module.
Furthermore, we encode the reference image using a VAE encoder and a CLIP image encoder. 
To ensure video consistency, we utilize two key modules: the ReferenceNet and the temporal alignment module. 
The VAE embeddings are fed into the ReferenceNet, which is responsible for maintaining consistency between the characters and background in the generated video and the reference image.
Additionally, we employ a motion alignment strategy that utilizes a series of motion modules to apply temporal attention across frames. 
This process aims to mitigate any discrepancies between the reference image and the motion guidance, thus enhancing the overall coherence of the generated video content.

\textbf{Training.}
The training process consists of two distinct stages.
During the initial stage, training is conducted solely on images, with the exclusion of motion modules within the model. 
We freeze the weights of the VAE encoder and decoder, as well as the CLIP image encoder, in a frozen state, while allowing the Guidance Encoder, Denoising U-Net, and reference encoder to be updated during training.
To initiate this stage, a frame is randomly selected from a human video to serve as a reference, and another image from the same video is chosen as the target image. 
The multi-layer guidance extracted from the target image is then input into the Guidance network. 
The primary objective of this stage is to generate a high-quality animated image utilizing the multilevel guidance derived from the specific target image.

In the second training phase, the incorporation of the motion module serves to augment the temporal coherence and fluidity of the model.
This module is initialized with the pre-existing weights obtained from AnimateDiff. 
A video segment comprising 24 frames is extracted and employed as the input data. 
During the training of the motion module, the Guidance Encoder, Denoising U-Net, and reference encoder, which were previously trained in the initial stage, are held constant.

\textbf{Inference.}
During the inference process, animation is performed on a specific reference image by aligning the motion sequences extracted from in-the-wild videos or synthesized ones.
Parametric shape alignment is utilized to align the motion sequence with the reconstructed SMPL model derived from the reference image at the pixel level, providing a basis for animation. 
To accommodate the input of a video clip comprising 24 frames, a temporal aggregation technique~\cite{tseng2022edge} is employed to concatenate multiple clips. This aggregation method aims to produce a long-duration video output.

\section{Experiments}
\label{sec:exp}

\subsection{Implementations}
\textbf{Dataset.}
We have curated an in-the-wild dataset comprising approximately 5,000 high-fidelity authentic human videos sourced from reputable online repositories, encompassing a total of 1 million frames. 
The dataset is segmented as follows: Bilibili (2540 videos), Kuaishou (920 videos), Tiktok \& Youtube (1438 videos), and Xiaohongshu (430 videos). 
These videos feature individuals of varying ages, ethnicities, and genders, depicted in full-body, half-body, and close-up shots, set against diverse indoor and outdoor backdrops.
In order to enhance our model's capacity to analyze a wide range of human movements and attire, we have included footage of dancers showcasing various dance forms in diverse clothing styles. 
In contrast to existing datasets characterized by pristine backgrounds, our dataset capitalizes on the diversity and complexity of backgrounds to aid our model in effectively distinguishing the foreground human subjects from their surroundings.
To maintain fairness and align with established benchmarks in the field of image animation, the identical test set as utilized in MagicAnimate~\cite{xu2023magicanimate} has been employed for TikTok evaluation.

\begin{table}[!t]
\centering
\begin{tabular}{c|cccc|ccc}
\hline
Method          & L1 $\downarrow$ & PSNR $\uparrow$ & SSIM $\uparrow$ & LPIPS $\downarrow$  & FID-VID $\downarrow$ & FVD $\downarrow$ \\ \hline
MRAA  & 3.21E-04        & 29.39           & 0.672           & 0.296              & 54.47                           & 284.82           \\
DisCo             & 3.78E-04            & 29.03             & 0.668             & 0.292                            & 59.90                  & 292.80              \\
MagicAnimate  & 3.13E-04    & 29.16           & 0.714           & 0.239              & 21.75                          & 179.07           \\
Animate Anyone  & -            & 29.56             & 0.718             & 0.285                & -                                & 171.9              \\
Ours & 3.02E-04            & 29.84             & 0.773            & 0.235                             & 26.14                  & 170.20   \\
Ours*  & \textbf{2.94E-04}            & \textbf{29.91}             & \textbf{0.802}            & \textbf{0.234}                             & \textbf{21.07}                  & \textbf{160.82}         \\\hline
\end{tabular} 
\vspace{1mm}
\caption{Quantitative comparisons on Tiktok dataset.
* indicates that the proposed approach is fine-tuned on the Tiktok training dataset.}
\vspace{-4mm}
\label{tab:quantitative_tiktok}
\end{table}

\begin{figure*}[!t]
  \centering
  \includegraphics[width=\textwidth]{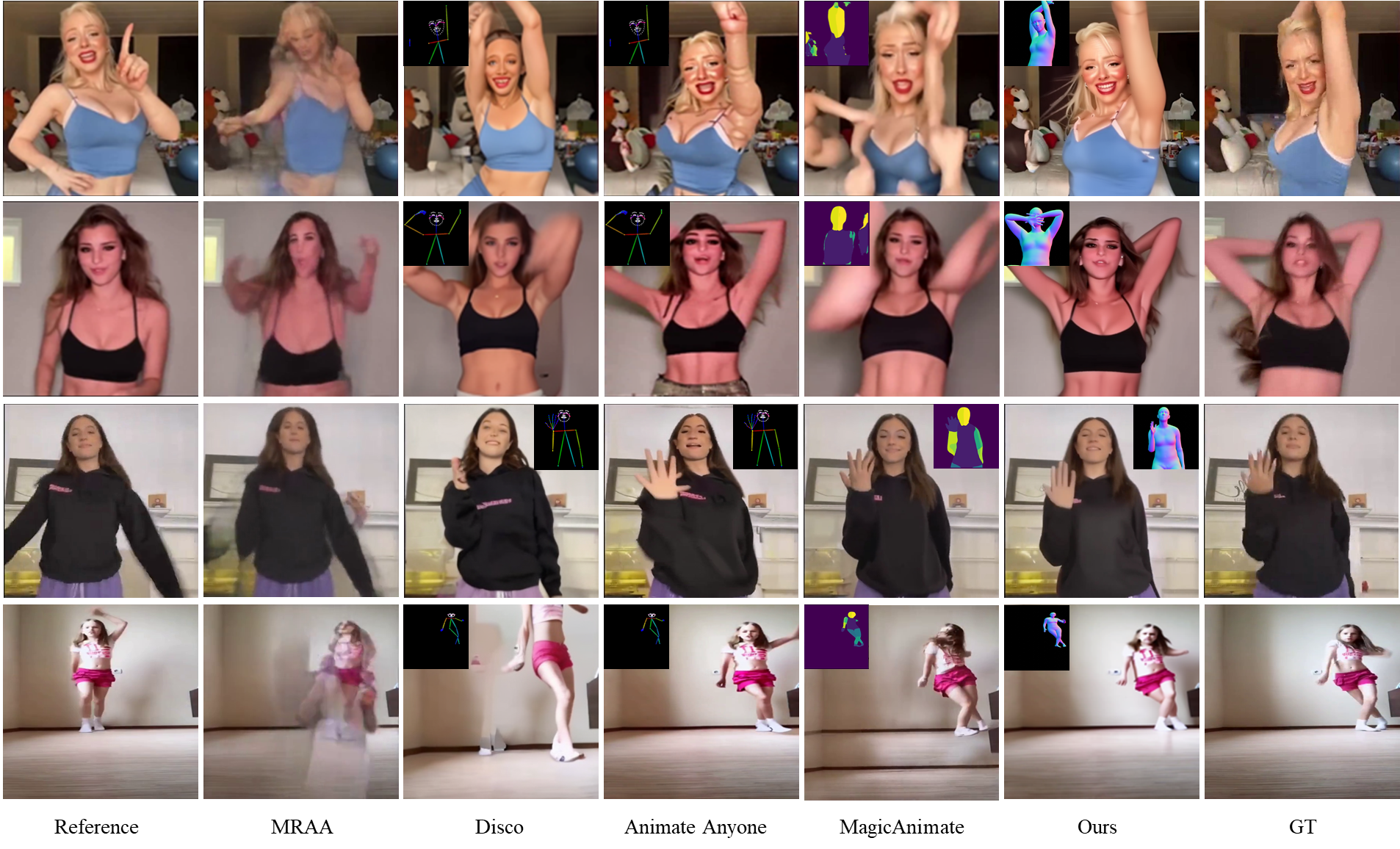}
  \caption[]{Qualitative comparisons between our and the state-of-the-art approaches on TikTok and proposed unseen dataset.}
  \vspace{-4mm}
  \label{fig:qualitative_comparisons}
\end{figure*}

\textbf{Implementation.} 
Our experiments were facilitated by the computational power of 8 NVIDIA A100 GPUs. 
The training regimen is structured in two phases: initially, we processed individual video frames—sampling, resizing, and center-cropping them to a uniform resolution of 768x768 pixels. 
This stage spanned 60,000 steps with a batch size of 32. Subsequently, we dedicated attention to the temporal layer, training it over 20,000 steps with sequences of 24 frames and a reduced batch size of 8 to enhance temporal coherence. Both stages adhered to a learning rate of 1e-5.
During inference, to achieve continuity over extended sequences, we employed a temporal aggregation method, which facilitated the seamless integration of results from distinct batches, thereby generating longer video outputs. 

\begin{figure*}[!t]
  \centering
  \includegraphics[width=\textwidth]{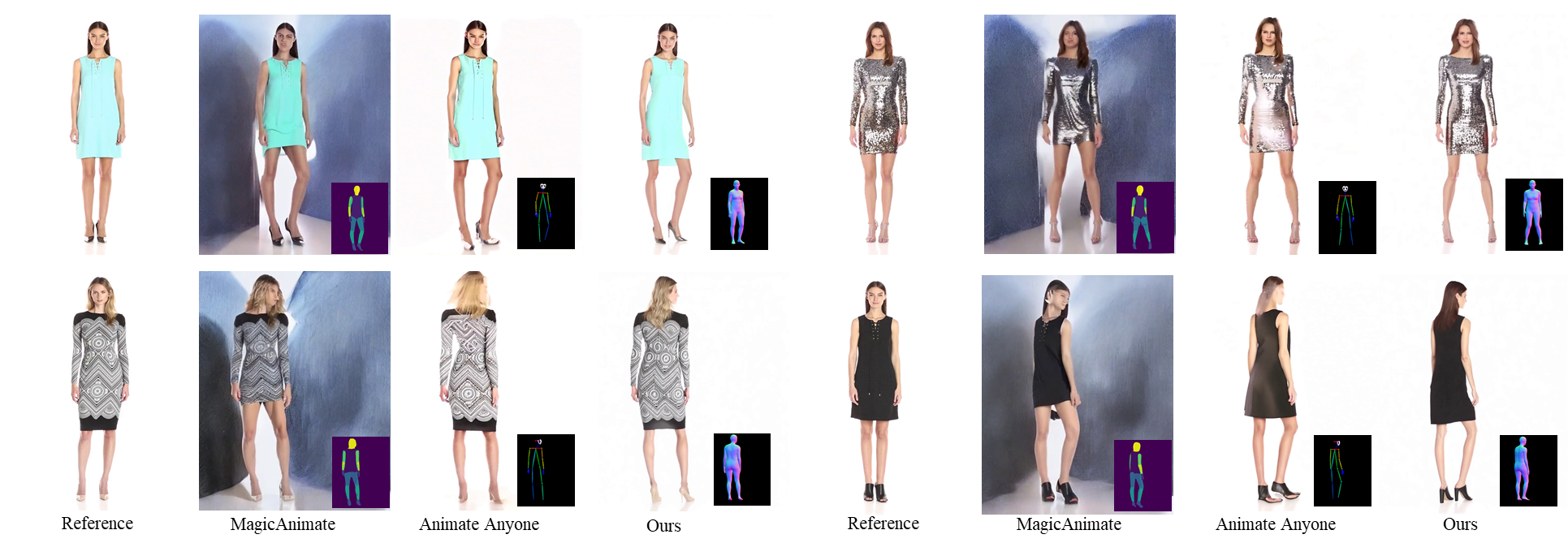}
  \caption[]{Qualitative comparisons between our and the state-of-the-art approaches on UBC fashion video datasets.}
  \label{fig:ubc_data}
\end{figure*}

\begin{figure*}[!t]
  \centering
  \includegraphics[width=\textwidth]{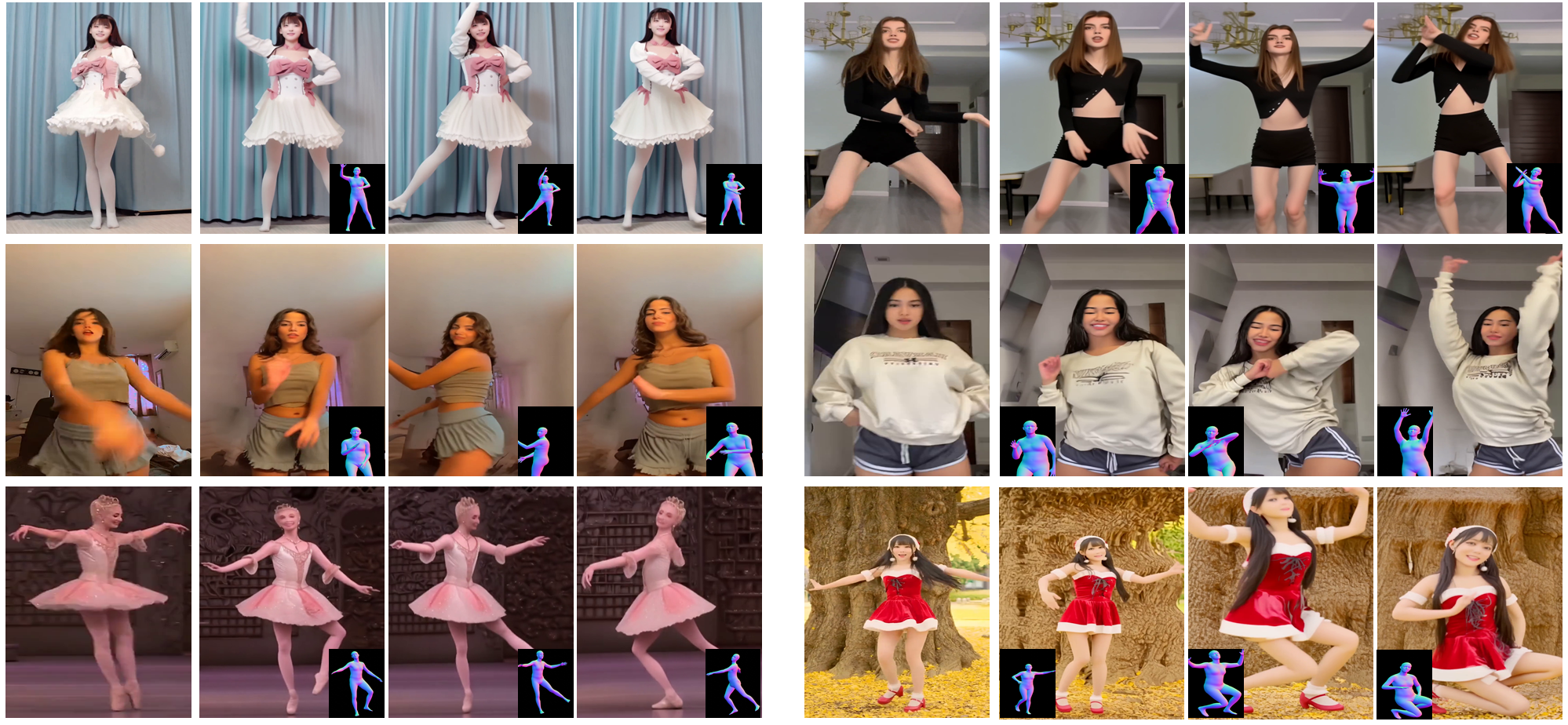}
  \caption[]{More qualitative results of our approach on the proposed unseen dataset.}
  \vspace{-4mm}
  \label{fig:wild_data}
\end{figure*}

\subsection{Comparisons}
\textbf{Baselines.} We perform a comprehensive comparison with several state-of-the-art methods for human image animation: (1) MRAA~\cite{siarohin2021motion} is state-of-the-art GAN-based animation approaches, which estimate optical flow from driving sequences to warp the source image and then inpaint the occluded regions using GAN models. (2) DisCo~\cite{wang2023disco} is the state-of-the-art diffusion-based animation method that integrates disentangled condition modules for pose, human, and background into a pretrained diffusion model to implement human image animation. (3) MagicAnimate~\cite{xu2023magicanimate} and Animate Anyone~\cite{hu2023animate} are newer diffusion-based animation methods that employ more complex model structure and train on more general data which makes them perform quite well on TikTok dataset.
In all the qualitative (video and visual) comparative experiments, we employed the open-source implementation of Animate Anyone from MooreThreads\footnote{\tiny Animate Anyone: \url{https://github.com/MooreThreads/Moore-AnimateAnyone}} and MagicAnimate from the original authors\footnote{\tiny MagicAnimate: \url{https://github.com/magic-research/magic-animate}}. 
For all quantitative experimental comparisons, we directly referenced the relevant statistics from the original literature.

\begin{table}[!t]
\centering
\begin{tabular}{c|cccc}
    \hline
    Method          & PSNR $\uparrow$ & SSIM $\uparrow$ & LPIPS $\downarrow$ & FVD $\downarrow$ \\ \hline
    MRAA    & -          & 0.663           & 0.311           & 321.5                                   \\
    DisCo & 28.32          & 0.694           & 0.286          & 295.4    \\ 
    MagicAnimate & 29.14          & 0.713           & 0.235          & 193.6    \\ 
    Animate Anyone  & 28.86  &  0.727  & 0.242 & 199.2\\
    Ours  & \textbf{29.87}          & \textbf{0.806}           & \textbf{0.211}          & \textbf{173.5}\\ \hline
\end{tabular} 
\vspace{1mm}
\caption{Quantitative comparisons on proposed unseen dataset.}
\vspace{-4mm}
\label{tab:quantitative_wild}
\end{table}

\begin{table}[!t]
\centering
\begin{tabular}{c|cccc|cc}
\hline
Method          & L1 $\downarrow$ & PSNR $\uparrow$ & SSIM $\uparrow$ & LPIPS $\downarrow$  & FID-VID $\downarrow$ & FVD $\downarrow$ \\ \hline
Ours (w/o. SMPL)  & 4.83E-04        & 28.57           & 0.672           & 0.296                  & 30.06                & 192.34          \\

Ours (w/o. geo.)  & 4.06E-04    & 28.78          & 0.714           & 0.276                        & 29.75                & 189.07           \\
Ours (w/o. skl.) & 3.76E-04            & 29.05            & 0.724             & 0.264            & 34.12                               & 184.24 \\
Ours & 3.02E-04          & 29.84           & 0.773            & 0.235          &  26.14                  & 170.20            \\\hline
\end{tabular}
\vspace{1mm}
\caption{Ablation study on different motion guidance.
``w/o. SMPL'' denotes a scenario where only the skeleton map is utilized as the motion condition.
``w/o. geo.'' indicates the model configuration that disregards geometric information, specifically depth and normal maps, as components of the motion condition.
``w/o. skl.'' describes the condition where the model solely relies on SMPL-derived inputs (including depth, normal, and semantic maps) for motion guidance.
``ours'' signifies the proposed approach that integrates both SMPL and skeleton derived motion condition.
}
\vspace{-4mm}
\label{tab:guidance_ablation}
\end{table}

\textbf{Evaluation metrics.} 
Our evaluation methodology adheres to the established metrics utilized in existing research literature.
Specifically, we assess both single-frame image quality and video fidelity. 
The evaluation of single-frame quality incorporates metrics such as the L1 error, Structural Similarity Index (SSIM)~\cite{wang2004image}, Learned Perceptual Image Patch Similarity (LPIPS)~\cite{zhang2018unreasonable}, and Peak Signal-to-Noise Ratio (PSNR)~\cite{5596999}. 
Video fidelity, on the other hand, is evaluated through the Frechet Inception Distance with Fréchet Video Distance (FID-FVD)~\cite{ijcai2019p276} and Fréchet Video Distance (FVD)~\cite{unterthiner2018towards}.

\textbf{Evaluation on benchmark dataset.}
Table~\ref{tab:quantitative_tiktok} presents a concise quantitative analysis of various methods evaluated on the TikTok dataset, focusing on key metrics such as L1 loss, PSNR, SSIM, LPIPS, FID-VID, and FVD. 
The proposed method, both in its original and fine-tuned (* indicated) forms, demonstrates superior performance across most metrics, particularly highlighting its effectiveness in achieving lower L1 loss, higher PSNR and SSIM values, and reduced LPIPS, FID-VID, and FVD scores. 
Notably, the fine-tuned version of our approach shows the best overall results, indicating the benefits of dataset-specific optimization. 
Figure~\ref{fig:qualitative_comparisons} and Figure~\ref{fig:ubc_data} provides additional qualitative comparison on such benchmark.

\begin{figure}[!t]
  \centering
  \includegraphics[width=1.0\linewidth]{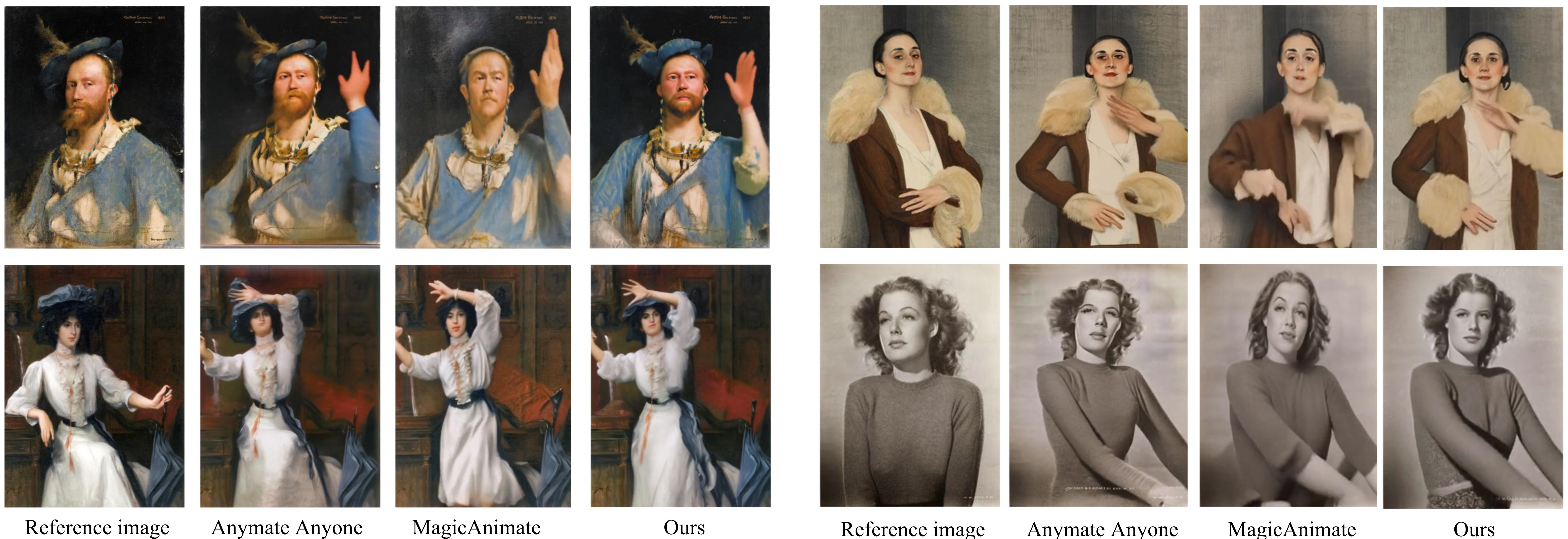}
  \caption{The qualitative comparison of animating unseen domain images.}
  \label{fig:unseen_data}
\end{figure}

\textbf{Evaluation on Proposed Unseen Dataset.}
In order to further compare the robustness of various methods, distinct from datasets such as TikTok and UBC fashion that exhibit domain proximity, we have constructed a testing dataset comprising 100 high-fidelity authentic human videos sourced from online repositories.
These videos exhibit significant variations in the shape, pose, and appearance of the individuals depicted.
Figure~\ref{fig:wild_data} and Figure~\ref{fig:unseen_data} provides some qualitative comparison of the unseen dataset along with the statistical comparison presented in Table~\ref{tab:quantitative_wild}, collectively illustrate the efficacy of the proposed approach in generalizing to unseen domains.

\begin{table}[!t]
\centering
\begin{tabular}{c|cccc|cccc}
\hline
Method          & L1 $\downarrow$ & PSNR $\uparrow$ & SSIM $\uparrow$ & LPIPS $\downarrow$ &  FID-VID $\downarrow$ & FVD $\downarrow$ \\ \hline
w/o.  & 3.21E-04        & 29.44           & 0.752           & 0.248                      & 25.36                & 174.46           \\
w/.  & 3.02E-04           & 29.84             & 0.785            & 0.235                              & 21.28                  & 170.20             \\\hline
\end{tabular} 
\vspace{1mm}
\caption{Ablation study on guidance self-attention.}
\vspace{-4mm}
\label{tab:guidance_attention}
\end{table}

\begin{figure}[!t]
  \centering
  \includegraphics[width=1.0\linewidth]{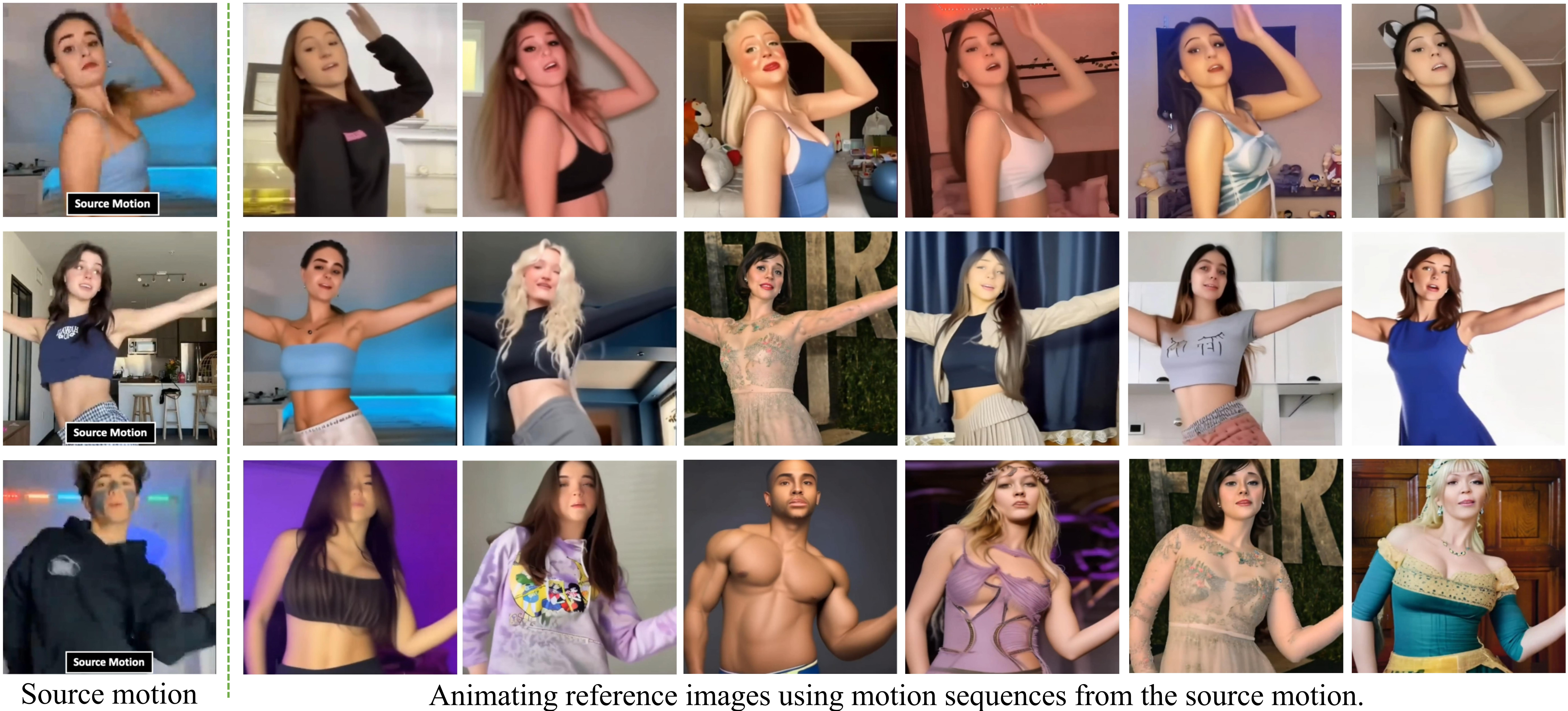}
  \caption{The demonstration of cross ID animation from the proposed approach.}
  \vspace{-4mm}
  \label{fig:cross_id}
\end{figure}

\begin{figure}[!t]
  \centering
  \includegraphics[width=1.0\linewidth]{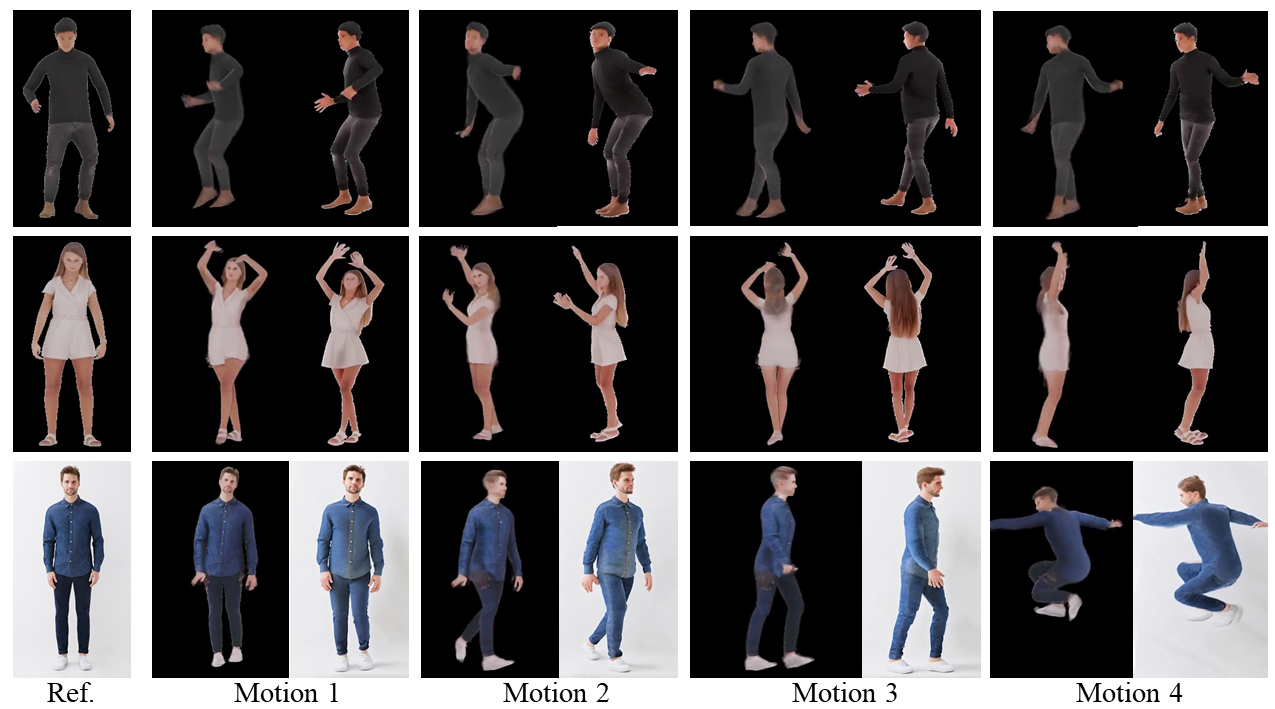}
  \caption{Comparision between SHERF (left) and ours (right).}
  \vspace{-4mm}
  \label{fig:comp_sherf}
\end{figure}

\begin{figure}[!t]
  \centering
  \includegraphics[width=0.8\linewidth]{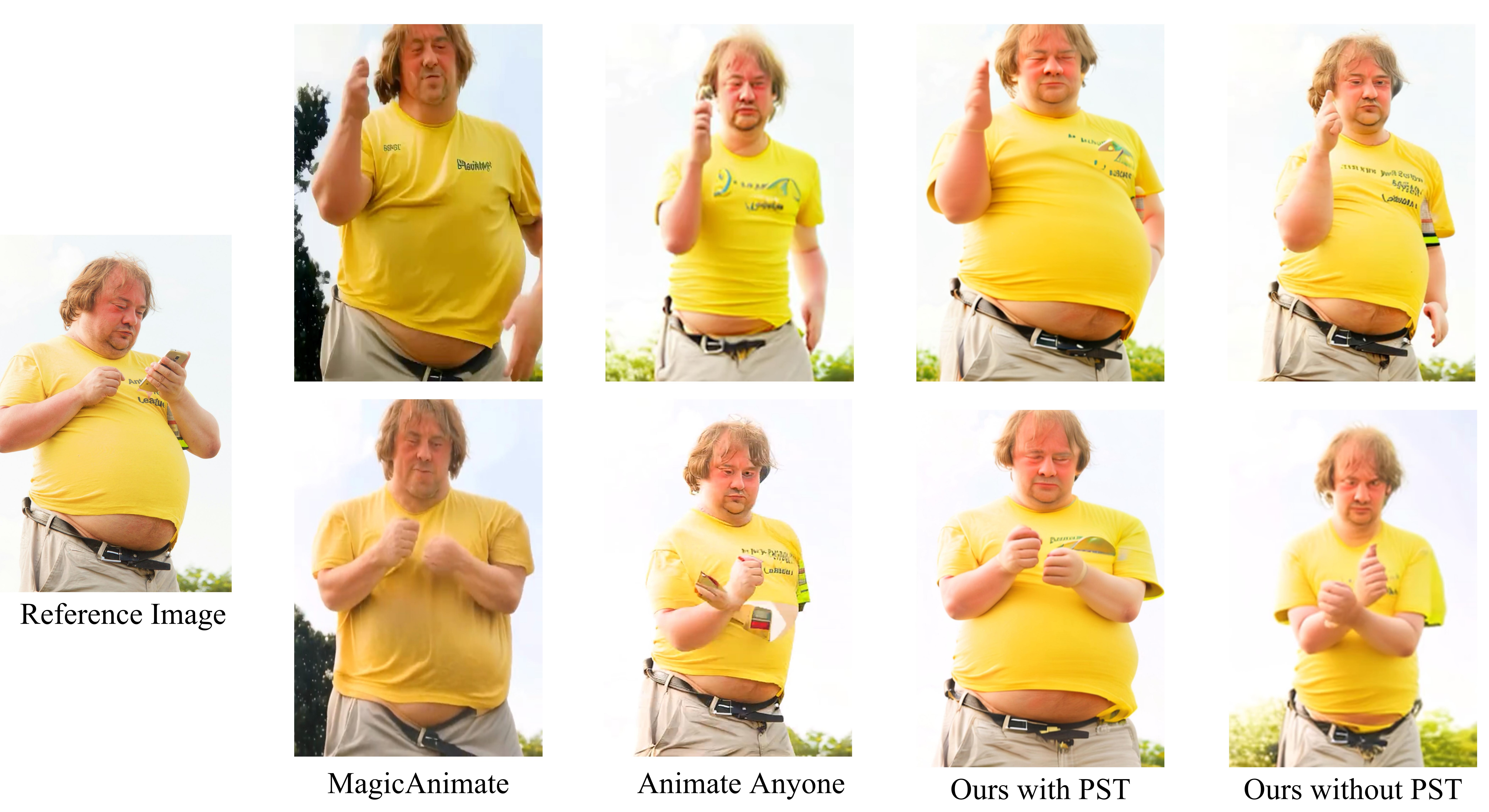}
  \caption{The comparison on the shape variance data.}
  \label{fig:shape_variance}
\vspace{-4mm}  
\end{figure}

\textbf{Cross ID Animation.}
As shown in Figure~\ref{fig:cross_id}, a comparative analysis is conducted between our approach and state-of-the-art baseline methods for cross-identity animation, specifically focusing on the task of animating reference images through motion sequences derived from disparate videos.

\textbf{Multi-view Animation.}
Although our method may not match the direct rendering of 3D human representations for consistent novel views, we employ a sequence of SMPLs for multi-view consistent motion guidance and utilize generative models to achieve satisfactory multi-view results.  As shown in Figure~\ref{fig:comp_sherf}, we compare the results of our multi-view animation with a representative single-image to 3D human reconstruction method, SHERF~\cite{hu2023sherf}.

\subsection{Ablation Studies}
\begin{figure}[!t]
  \begin{minipage}{0.48\textwidth}
     \centering
    \includegraphics[width=1.0\textwidth]{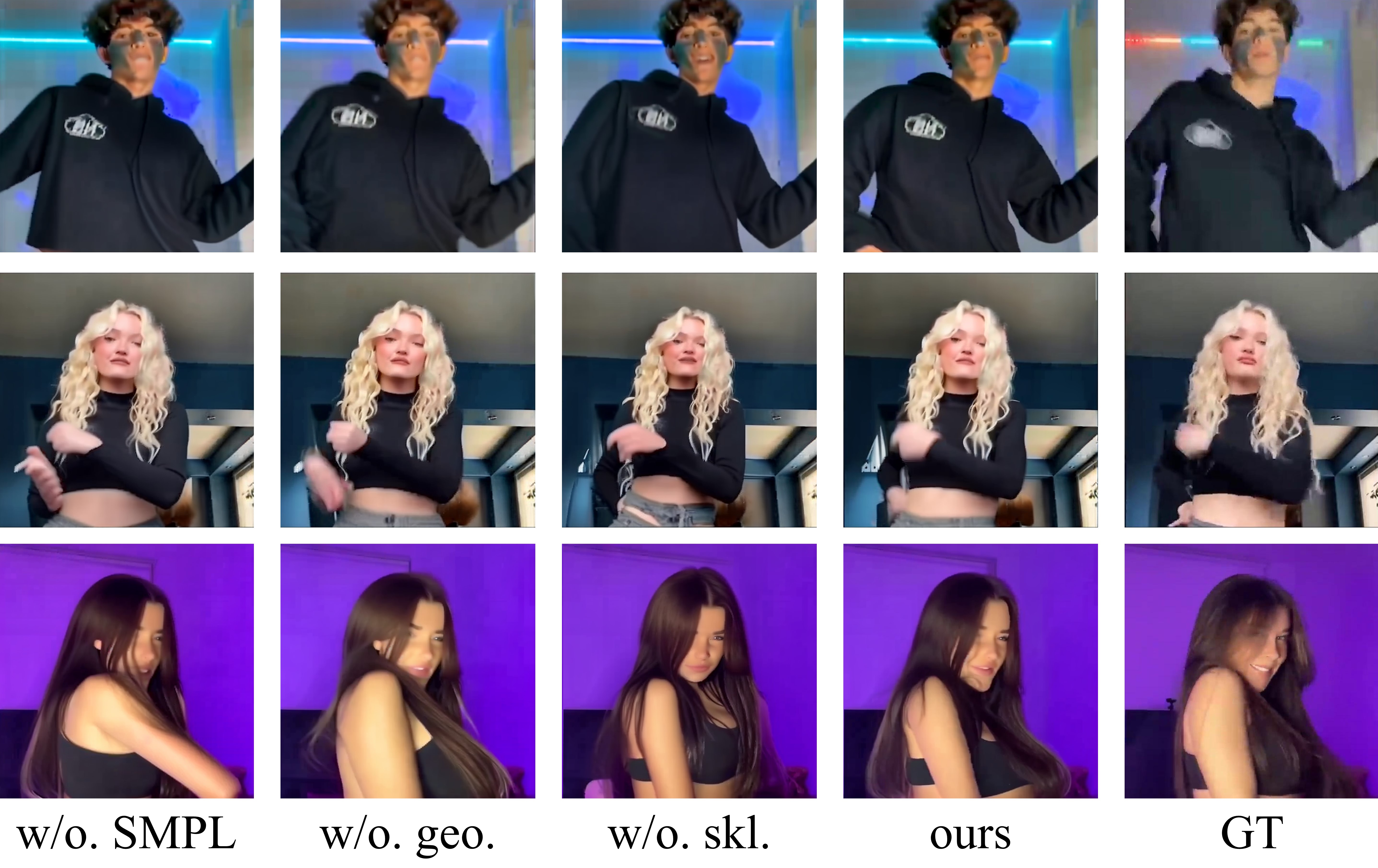}
    \caption{Ablation analysis on different motion conditions. geo. refers to the geometry. skl. is the skeleton condition.}
    \label{fig:ablation_motion}
    
  \end{minipage}
  \hfill
   \begin{minipage}{0.48\textwidth}
    \centering
    \includegraphics[width=0.9\textwidth]{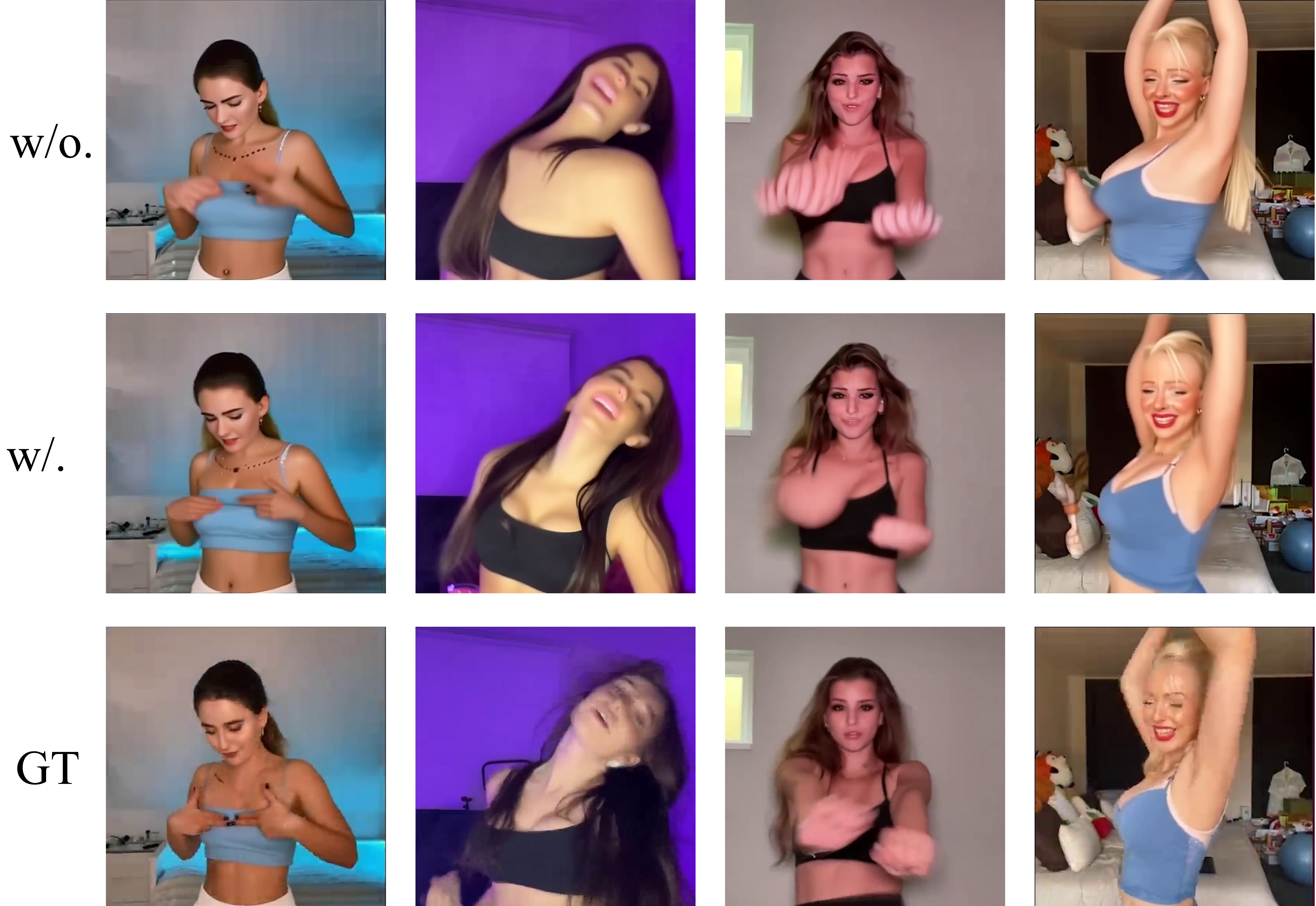}
    \caption{Effect of guidance attention. w/. and w/o. indicate the guidance with and without self-attention.}
    \label{fig:ablation_guidance}
  \end{minipage}
\end{figure}

\textbf{Different Conditions from SMPL.}
As shown in Table~\ref{tab:guidance_ablation}, the statistics demonstrate that the full configuration of the proposed method (``ours'') obviously outperforms other variants in terms of image quality and fidelity, and video consistency and realism. 
The ablation components, ``w/o SMPL'', ``w/o geometry'', and ``w/o skeleton'', show progressively improved performance as more components are included.
Specifically, SMPL obviously brings more gains in PSNR (1.27 v.s. 0.48) and SSIM (0.10 v.s. 0.05) gains than a skeleton, which means better preserving the shape alignment and motion guidance.
Moreover, it is noteworthy that the incorporation of both the SMPL and skeleton models leads to additional improvements. 
Specifically, the skeleton model exhibits advantages in providing refined motion guidance for facial and hand regions.
Meanwhile, Figure~\ref{fig:ablation_motion} qualitative demonstrates the effectiveness of different conditions from SMPL.

\textbf{Guidance Self-Attention.}
Table~\ref{tab:guidance_attention} presents the findings of an ablation study conducted on guidance self-attention. 
The results indicate that the inclusion of guidance attention leads to superior performance compared to the absence of such attention, as evidenced by improvements across all evaluated metrics. 
As shown in Figure~\ref{fig:ablation_guidance}, we provide additional qualitative results of the guidance self-attention.

\textbf{Parametric Shape Alignment.}
As shown in Figure~\ref{fig:shape_variance},  we take an ablation study on shape parameter alignment between reference humans' SMPL model and driving videos' SMPL sequence. To highlight the effect, we use an individual with an extreme figure as the reference image and a common human dancing video as input. In comparison to other methods, our results from parametric shape alignment in the third row exhibit the most consistent shape and figure alignment with the reference image.

\textbf{Efficiency Analysis.}
Table~\ref{tab:efficiency} presents an efficiency analysis of our proposed approach, detailing the GPU memory requirements and time consumption for different steps, including parametric shape transfer, rendering per frame, and inference per frame.

\begin{table}[!t]
\centering
\begin{tabular}{c|ccccc}
\hline
Method          & GPU memory (GB) & & & & Time (sec) \\ \hline
Parametric shape transfer    &   3.24     & & & & 0.06 \\
Rendering (per frame) &  2.86    & & & & 0.07 \\
Inference (per frame)  & 19.83       & & & & 0.52 \\\hline
\end{tabular} 
\vspace{1mm}
\caption{Efficiency analysis of different steps of the proposed approach.} 
\label{tab:efficiency}
\vspace{-4mm}  
\end{table}

\subsection{Limitations and Future Works}
Despite the enhanced shape alignment capabilities and motion guidance offered by the human parametric SMPL model, its modeling capacity for faces and hands is limited. 
Consequently, the guidance effect for faces and hands does not match the efficacy of feature-based methods. This limitation underpins the incorporation of DWpose as an additional constraint for facial and hand modeling. 
As illustrated in Figure~\ref{fig:motion_guidance}, the self-attention mechanism further amplifies the saliency of faces and hands within the skeleton map. 
However, it is important to note that since SMPL and DWpose are solved independently, a potential discrepancy in consistency between them exists. 
Although this discrepancy did not manifest significantly in our experiments, it nonetheless introduces a potential source of error.
\section{Conclusion}
\label{sec:conclusion}

This paper introduces a novel approach to human image animation that integrates the SMPL 3D parametric human model with latent diffusion models, aiming to enhance pose alignment and motion guidance. 
By leveraging the unified representation of shape and pose variations offered by the SMPL model, along with depth, normal, and semantic maps, this method further improve the ability of capturing realistic human movements and shapes of previous techniques.
The inclusion of skeleton-based motion guidance and self-attention mechanisms for feature map integration further refines the animation process, enabling the creation of dynamic visual content that more accurately reflects human anatomy and movement. 
Experimental validation across various datasets confirms the effectiveness of this approach in producing high-quality human animations, showcasing its potential to advance digital content creation in fields requiring detailed and realistic human representations.


%
%
\bibliographystyle{splncs04}
\bibliography{main}
\end{document}